# Token Adaptation via Side Graph Convolution for Efficient Fine-tuning of 3D Point Cloud Transformers

**Takahiko Furuya**[1]
[1]Integrated Graduate School of Medicine, Engineering, and Agricultural Sciences, University of Yamanashi, 4-3-11 Takeda, Kofu-shi, Yamanashi-ken, 400-8511, Japan

Corresponding author: Takahiko Furuya (e-mail: takahikof@yamanashi.ac.jp).

This work was supported by the Japan Society for the Promotion of Science (JSPS) KAKENHI (Grant No. 24K14992).

**ABSTRACT** Parameter-efficient fine-tuning (PEFT) of pre-trained 3D point cloud Transformers has emerged as a promising technique for 3D point cloud analysis. While existing PEFT methods attempt to minimize the number of tunable parameters, they often suffer from high temporal and spatial computational costs during fine-tuning. This paper proposes a novel PEFT algorithm called *Side Token Adaptation on a neighborhood Graph* (*STAG*) to achieve superior temporal and spatial efficiency. STAG employs a graph convolutional side network operating in parallel with a frozen backbone Transformer to adapt tokens to downstream tasks. Through efficient graph convolution, parameter sharing, and reduced gradient computation, STAG significantly reduces both temporal and spatial costs for fine-tuning. We also present *Point Cloud Classification 13* (*PCC13*), a new benchmark comprising diverse publicly available 3D point cloud datasets to facilitate comprehensive evaluation. Extensive experiments using multiple pre-trained models and PCC13 demonstrates the effectiveness of STAG. Specifically, STAG maintains classification accuracy comparable to existing methods while reducing tunable parameters to only 0.43M and achieving significant reductions in both computation time and memory consumption for fine-tuning. Code and benchmark will be available at: https://github.com/takahikof/STAG.

## 1. Introduction

Analyzing 3D point cloud data has become increasingly significant due to its wide-ranging application scenarios, including autonomous driving, robotics, infrastructure maintenance, and disaster prevention. Recent advancements in deep learning have led to the development of sophisticated 3D point cloud Transformers, particularly those leveraging the transfer learning framework [1, 2, 3]. Transfer learning involves pre-training a backbone deep neural network (DNN) on a large dataset, followed by fine-tuning the backbone DNN for a specific downstream task. Notably, self-supervised pre-training methods that can exploit unlabeled 3D point cloud datasets have been extensively studied under both single-modal setting [4, 5, 6] and cross-modal setting [7, 8, 9].

In contrast to the success of pre-training, research on fine-tuning pre-trained DNNs remains underexplored in the field of 3D point cloud analysis. The predominant fine-tuning approach, known as full fine-tuning, adjusts all parameters within the pre-trained backbone DNN. Despite its simplicity, full fine-tuning faces several limitations that hinder its practicality in real-world scenarios of 3D point cloud analysis. That is, full fine-tuning incurs significant storage costs since the tuned parameters must be stored separately for each downstream task. The storage cost becomes more serious as the scale of the backbone DNN increases. In addition, full fine-tuning is computationally inefficient as it requires calculating gradients for all parameters of the backbone during backpropagation. Computing gradients for all parameters results in increased memory consumption and longer training time. Furthermore, tuning all parameters is prone to overfitting and catastrophic forgetting, which potentially diminishes the generalization capability of the pre-trained backbone.

Recent research has focused on parameter-efficient fine-tuning (PEFT) of 3D point cloud Transformers [10–15]. This paper uses the abbreviation "PEFT-PT" to denote PEFT specifically designed for 3D point cloud Transformers. The existing PEFT-PT methods attempt to address the limitations of full fine-tuning, particularly in terms of storage cost. The PEFT-PT methods freeze most parameters of the pre-trained backbone Transformer and fine-tune only a small subset of parameters, either within the backbone or additional adaptation modules. PEFT-PT has significantly lowered the storage cost for fine-tuned parameters, while maintaining analysis accuracy comparable to full fine-tuning.

However, we argue that the existing studies on PEFT-PT still have three shortcomings from the perspectives of method and evaluation. First, the existing methods suffer from a long computation time and large memory consumption during fine-tuning. Such temporal and spatial inefficiency primarily stem from the adaptation modules used by the existing methods. Adaptation modules, typically designed as multi-layer perceptrons (MLPs), are often inserted not only in the deeper layers but also in the *shallower* layers of the Transformer backbone (Fig. 1a). Therefore, even when all the parameters within the backbone are frozen, the gradients of the frozen backbone must be computed during backpropagation. In addition, most existing methods generate additional tokens, or feature vectors processed by Transformer, to effectively adapt the frozen backbone to downstream tasks. However, increasing the tokens also incurs slow training and large memory footprint.

The second issue is difficulty of implementation. Most PEFT-PT methods modify the internal architecture of the Transformer backbone. Given that different 3D point cloud Transformers have varying internal architectures, applying the existing PEFT-PT methods to a new Transformer is not straightforward. That is, it requires a deep understanding of both the PEFT-PT method and the backbone architecture being used. This implementation cost serves as an obstacle to the adoption of the existing PEFT-PT methods and limits their versatility.

The third issue lies in the evaluation process. Existing studies on PEFT-PT rely on a limited number of datasets, i.e., ScanObjectNN [16] and ModelNet [17] only, to evaluate downstream task performance. Such a small-scale evaluation hinders our understanding of the generalizability and robustness of PEFT-PT methods across a wide range of point cloud datasets. Moreover, designing methods based on a small number of datasets risks over-adaptation to these specific benchmarks, potentially compromising adaptability to diverse point cloud datasets.

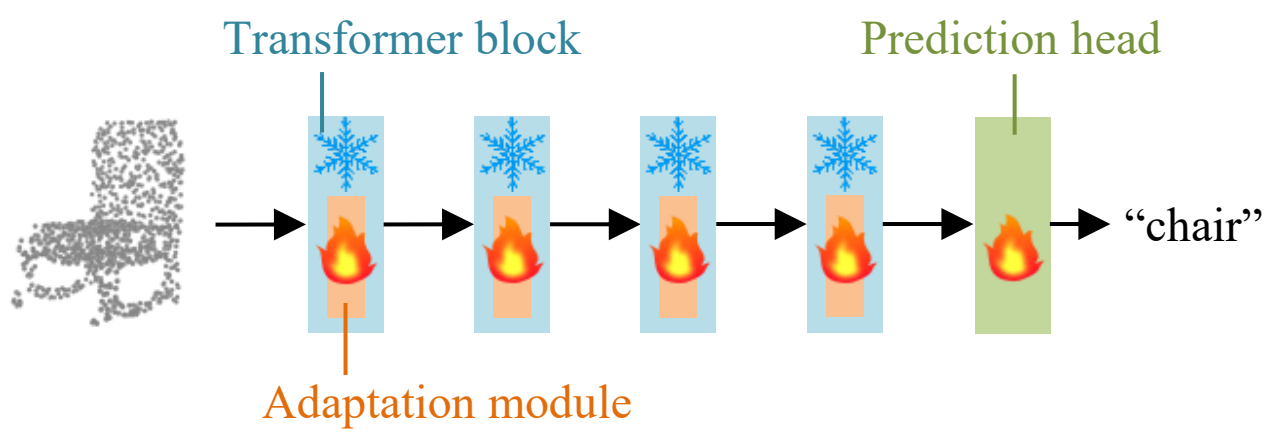

(a) Existing PEFT-PT approach

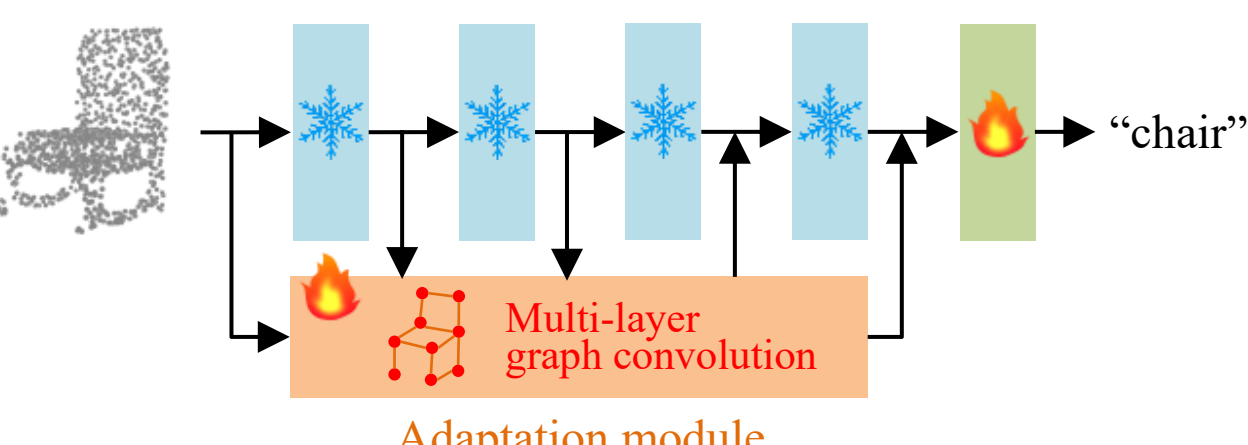

(b) Proposed PEFT-PT approach

**Fig. 1** Difference between conventional and proposed PEFT-PT approaches. (a) Existing methods insert adaptation modules inside each Transformer block, which requires gradient computations for all the blocks. (b) The proposed method employs a side graph convolutional network that is partially independent of the backbone, thereby reducing the gradient computation for the Transformer blocks. In the example of the figure, the gradient computation for the first three blocks can be omitted.

The abovementioned shortcomings motivate us to achieve the following two goals. (1) Developing a PEFT-PT algorithm that is temporally and spatially efficient as well as versatile. (2) Establishing a new benchmark that enables us to evaluate the generalizability of PEFT-PT methods across diverse 3D point cloud datasets.

To achieve the first goal, we propose a novel PEFT-PT algorithm called *Side Token Adaptation on a neighborhood Graph* (*STAG*). Unlike conventional methods that insert adaptation modules inside the Transformer backbone, STAG employs an adaptation module that runs in parallel with the backbone (Fig. 1b). The core operation of STAG is a graph convolution applied to a spatial neighborhood graph. Graph convolution is widely recognized as a powerful and universal operation capable of extracting hierarchical 3D shape features [18, 19, 73]. We expect that graph convolution helps tokens effectively adapt to downstream tasks via feature refinement considering the spatial relations among tokens. To enhance efficiency, STAG incorporates three improvements. First, the fusion of the tokens processed by the backbone with those processed by the adaptation module takes place only in the latter part of the backbone (Fig. 1b). Such a restricted connection reduces computation overhead for backpropagation. Second, a parameter sharing framework is introduced across multiple layers within the adaptation module to reduce parameter redundancy. Third, we employ the popular graph convolution operator EdgeConv [20] but modify it to improve its efficiency.

The proposed algorithm has two major advantages. Firstly, STAG is fast and memory-efficient. Since gradient computation is required only for the later layers of the backbone, the computational cost of backpropagation can be significantly reduced. In addition, parameter sharing across adaptation modules allows for a substantial reduction in the number of tunable parameters. Second, STAG is easy to implement and highly versatile. That is, STAG performs token adaptation independently of the processing of Transformer blocks. Therefore, STAG does not require modifications to the internal architecture of Transformer, making it easy to apply STAG to various 3D point cloud Transformers.

To achieve the second goal, this paper proposes a new classification benchmark called *Point Cloud Classification 13* (*PCC13*). PCC13 consists of 13 publicly available labeled 3D point cloud datasets. The datasets vary in terms of scale, point cloud type (synthetic or realistic), and category distribution. PCC13 thus enables the evaluation of generalizability of the PEFT-PT methods across diverse 3D point cloud data.

The comprehensive evaluation using the PCC13 benchmark and multiple pre-trained 3D point cloud Transformers demonstrates the efficacy of the STAG algorithm. STAG exhibits competitive or superior accuracy to the existing PEFT-PT methods, while achieving higher efficiency. Notably, fine-tuning by STAG requires only 0.43M tunable parameters, operating 1.4 times faster than DAPT [11], and reducing VRAM consumption by 40% compared to PointGST [15].

Contributions of this paper can be summarized as follows.

- Proposing a temporally and spatially efficient PEFT-PT algorithm called STAG. STAG leverages side graph convolutional network placed alongside the backbone, ensuring high accuracy, efficiency, and versatility.
- Proposing a new evaluation benchmark called PCC13. PCC13 facilitates robust evaluation of PEFT-PT methods across various 3D point cloud datasets.
- Evaluating STAG using PCC13. We validate that STAG exhibits notable overall performance in terms of both accuracy and efficiency.

The remainder of this paper is organized as follows. Related studies are reviewed in Section 2. Section 3 elaborates on the proposed algorithm and benchmark, followed by experimental evaluations in Section 4. Section 5 concludes this paper and discusses future work.

## 2. Related work

### 2.1 Pretraining of 3D point cloud Transformers

Transformer [21] has become an essential backbone DNN in various fields including 3D point cloud analysis [22]. Recently, self-supervised learning (SSL) of 3D point cloud Transformers has garnered significant attention [1–3, 23]. SSL leverages unlabeled 3D point clouds, thereby eliminating the need for laborious annotation. By solving a carefully designed pretext task, the backbone DNN acquires the ability to extract meaningful shape features. After pre-training by SSL, the backbone undergoes supervised fine-tuning to adapt to various downstream tasks. The pre-training of 3D point cloud Transformers can be roughly categorized into two settings: single-modal and multi-modal.

Single-modal SSL, which relies solely on 3D point cloud data, has two mainstream pretext tasks, i.e., shape reconstruction and feature contrast. Among the pretext tasks of shape reconstruction, masked autoencoding [4–6, 24–28] has emerged as a promising approach. Masked autoencoding randomly mask multiple local regions of the input point cloud, and the Transformer is tasked with reconstructing the masked regions. Such a pretext task helps the backbone learn generalizable feature representations that are robust against partial occlusion and missing parts. The other major pretext task, i.e., feature contrast [29–36, 74], leverages the framework of self-supervised contrastive learning [37]. In this paradigm, the backbone DNN is trained by optimizing a distance metric among the latent features of 3D point clouds. The goal of feature contrast is to bring the latent features of positive pairs closer together while pushing apart those of negative pairs. The training pairs can be formed at various levels, such as point-level, object-level, or scene-level, without relying on semantic labels.

Multi-modal SSL incorporates additional data modalities such as 2D images or text documents alongside 3D point clouds. The contrastive learning framework [37] is widely adopted to form a latent feature space shared across different modalities. A notable example is bimodal contrastive learning, which combines 3D point clouds and 2D images [38–45]. In this method, each positive pair is created between a point cloud and an image derived from the same object/scene, while a negative pair is formed by using different objects/scenes. By leveraging visual features that cannot be captured by 3D point clouds alone, the bimodal SSL facilitates the formation of a latent feature space more accurate than single-modal SSL. Some studies [7–9, 46–48] have explored trimodal contrastive learning that utilizes 3D point clouds, 2D images, and texts. This approach leverages powerful vision-language models such as CLIP [49], which is pre-trained via image-text feature contrast. Training data for trimodal contrastive learning are triplets, each consisting of a 3D point cloud, a 2D image, and a text description. By embedding 3D point cloud features into CLIP's latent space, the point cloud Transformer develops the ability to extract highly semantic shape features.

All the pre-trained 3D point cloud Transformers mentioned in this section undergo full fine-tuning to adapt to downstream tasks. However, as mentioned in Section 1, full fine-tuning faces challenges such as overfitting and high computational costs.

### 2.2 Parameter-efficient fine-tuning (PEFT)

#### 2.2.1 PEFT for vision/language Transformers

PEFT has emerged as a promising alternative to full fine-tuning, particularly in the fields of vision and language [50–52]. Despite the diversity of PEFT techniques, they share a common objective, i.e., adapting token features to a specific downstream task by tuning a limited number of parameters. PEFT mitigates the risk of overfitting and reduces the storage cost for the fine-tuned parameters. This subsection reviews three PEFT approaches related to this paper: adapter tuning, prompt tuning, and side tuning.

Adapter tuning (e.g., [53–55, 72]) incorporates adaptation modules into either the self-attention layer [54, 55, 72] or its subsequent MLP layer [53] within Transformer blocks. During fine-tuning, the parameters of the adaptation modules are adjusted while freezing the parameters of the backbone Transformer. Prompt tuning (e.g., [56, 57]) appends task-specific learnable prompts to the sequence of input tokens instead of adding adaptation modules. The token features are processed by the frozen Transformer and adapt to a downstream task through interaction with the task-specific prompts. Side tuning (e.g., [58–61]) employs a small auxiliary network that operates in parallel with the frozen backbone. The token features extracted by the backbone are combined with those extracted by the side network, typically through summation, to adapt to a downstream task.

From a perspective of complexity, side tuning is more efficient than adapter tuning and prompt tuning [60, 61]. In both adapter tuning and prompt tuning, tunable parameters exist upstream of the frozen backbone, necessitating gradient computation for nearly all neuron activations in the backbone. This requirement increases computation time and GPU memory consumption during fine-tuning. In contrast, side tuning can reduce gradient calculations for backbone parameters since the operation of the side network is entirely or partially independent of the backbone. Our proposed algorithm, i.e., STAG, adopts the side tuning approach to realize efficient fine-tuning.

#### 2.2.2 PEFT for 3D point cloud Transformers

Compared to vision/language Transformers, PEFT for 3D point cloud Transformers (PEFT-PT) has not been sufficiently explored. Existing PEFT-PT methods [10–15] adopt hybrid strategies combining adapter tuning and prompt tuning. For example, IDPT [10] generates an instance-aware dynamic prompt by using an adaptation module inserted into the penultimate Transformer block. IDPT fine-tunes the adaptation module in addition to the classification (CLS) token, which is located at the most upstream of the Transformer. DAPT [11] dynamically generates additional prompts by using sub-networks inserted into each Transformer block. In addition, DAPT incorporates adaptation modules into each Transformer block to adjust the distribution of token features by scaling and translation. Point-PEFT [12] and PPT [13] dynamically generate multiple prompts prior to input to Transformer and utilize adaptation modules inserted within each Transformer block. Adapter-X [14] generates a dynamic prompt and employs adaptation modules with a mixture-of-experts mechanism. PointGST [15] inserts adaptation modules that perform graph spectral analysis within each Transformer block and fine-tunes these adaptation modules and the CLS token.

While these PEFT-PT methods have successfully adapted to downstream tasks with a limited number of fine-tuned parameters, they have several drawbacks, as mentioned in Section 1. That is, the presence of tunable parameters upstream in the backbone increases the computational cost of gradient calculations. In addition, generating additional prompts leads to the computational overhead of Transformer. Furthermore, the tight coupling between adaptation modules and the backbone Transformer poses implementation challenges. From an evaluation perspective, existing studies have primarily validated their methods on only two datasets, i.e., ScanObjectNN and ModelNet, leaving their effectiveness on other point cloud datasets unclear.

This paper adopts an approach different from the existing methods, i.e., side tuning, to improve efficiency and ease of implementation. Moreover, we propose a new benchmark combining diverse 3D point cloud datasets to enable a comprehensive comparison of PEFT-PT methods.

## 3. Proposed algorithm and benchmark

### 3.1 Proposed algorithm: STAG

#### 3.1.1 Overview of STAG

Fig. 2 illustrates the processing pipeline of STAG. The adaptation module of STAG is a lightweight side network, which is partially independent of the backbone Transformer. The adaptation module comprises two components: Accumulation blocks (A-blocks) and Modulation blocks (M-blocks). A-blocks, positioned in the earlier part of the adaptation module, accumulate tokens extracted by the tokenizer and each Transformer block. On the other hand, M-blocks, placed in the latter part, not only accumulate tokens but also refine them by using graph convolution. The refined tokens are then fed back into the latter blocks of the backbone. During fine-tuning, only the parameters of the adaptation module and prediction head are updated, while those of Tokenizer and backbone remain frozen.

We incorporate three improvements into STAG to enhance its efficiency. First, A-blocks are designed to have a unidirectional data flow. That is, each A-block receives tokens from the backbone and transmits them to a subsequent A-block. Such layer connections eliminate the need for gradient computation in the earlier Transformer blocks. Second, sharing parameters among the same layer type within the adaptation module significantly reduces the number of tunable parameters. Third, we employ the powerful graph convolution operator, i.e., EdgeConv [20], but modify it to improve efficiency during fine-tuning.

**Notations:** In general, 3D point cloud Transformers process an input point cloud as a set of $n$ tokens, each of which representing a local region sampled from the input. We denote the token set by $\mathrm{T}^l=\{\mathbf{t}_i^l \in \mathbb{R}^d \mid i \in \{1, 2, \ldots, n\}\}$. The index $l \in \{0, 1, \ldots, L\}$ represents the position within the backbone, which consists of $L$ Transformer blocks. $\mathrm{T}^0$ is the initial tokens, i.e., the output from the Tokenizer, and $\mathrm{T}^L$ means the output from the final Transformer block. Each token is associated with the patch center coordinates. The set of patch centers is denoted as $\mathrm{C}=\{\mathbf{c}_i \in \mathbb{R}^3 \mid i \in \{1, 2, \ldots, n\}\}$. Typical 3D point cloud Transformers use the settings $n$=64 to 128, $d$=384, and $L$=12. Additionally, we use $\mathrm{X}^l=\{\mathbf{x}_i^l \mid i \in \{1, 2, \ldots, n\}\}$ to denote the output from the $l$-th adaptation block of STAG. The hyperparameter $A$ controls the number of A-blocks. That is, the first $A$ adaptation blocks constitute A-blocks, while the subsequent $L-A$ blocks belong to M-blocks.

Algorithm 1 presents the pseudocode of STAG. The lines colored in green are the STAG-specific processing. Omitting these lines boils down to the original point cloud Transformer. Note that STAG does not change the processing within the Transformer block. This design approach makes STAG easier to implement than the conventional PEFT-PT methods.

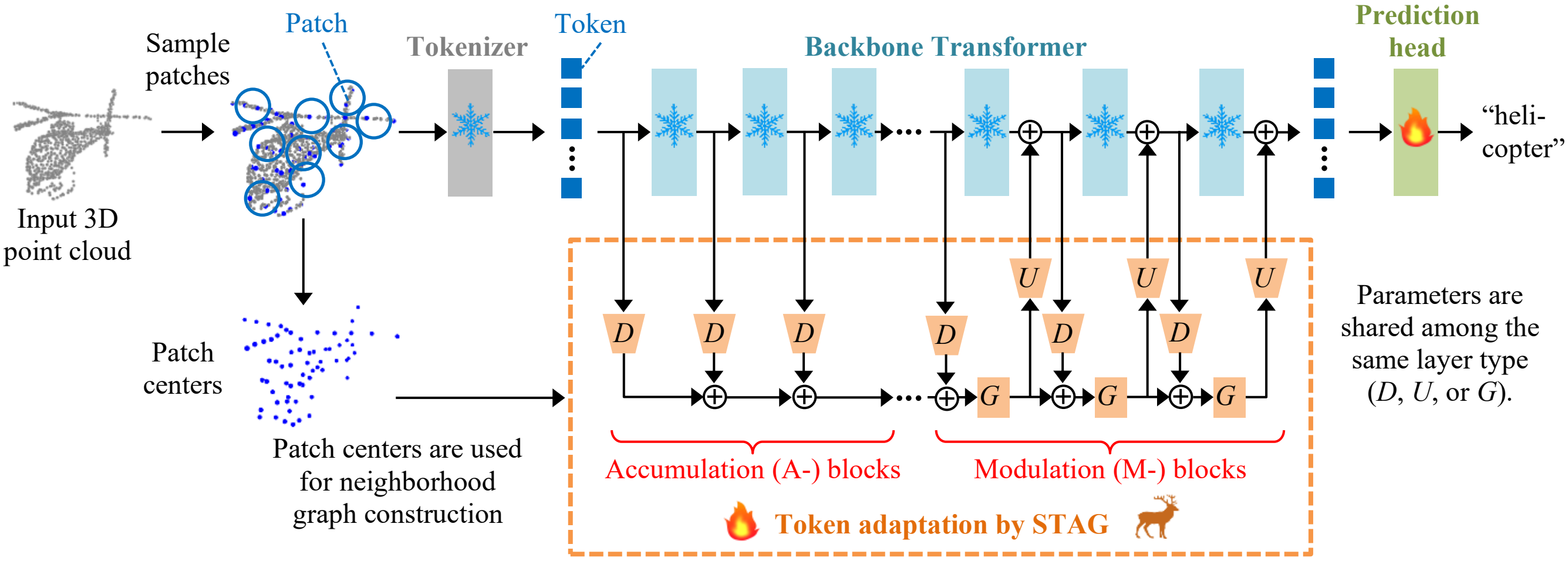

**Fig. 2** Overview of the proposed PEFT-PT algorithm, i.e., STAG, which employs a graph convolutional side network for efficient fine-tuning. The first half of STAG, called Accumulation blocks (A-blocks), receives and aggregates the tokens from the backbone. The latter half, called Modulation blocks (M-blocks), applies efficient graph convolution to the tokens and send them back to the backbone. Such layer connections omit the gradient computation in the first half of the backbone, thereby enabling efficient fine-tuning. STAG also employs parameter-sharing among each layer type, i.e., down projection ($D$), up projection ($U$), and graph convolution ($G$), to enhance parameter-efficiency.

### 3.1.2 Accumulation block (A-block)

The A-blocks simply accumulate the tokens extracted by the frozen Transformer blocks. Each input token $\mathbf{t}_i^{l-1}$ is processed by the following equation.

$$\mathbf{x}_i^l = D(\mathbf{t}_i^{l-1}) + \mathbf{x}_i^{l-1} \tag{1}$$

**Algorithm 1** Token adaptation by STAG

1: **Inputs:** Initial tokens $\mathrm{T}^0$
2: Patch centers C
3: Number of A-blocks $A$
4: **Output:** Adapted tokens $\mathrm{T}^L$
5: Initialize $\mathrm{X}^0$ by zero-vectors
6: **for** $l = 1, \ldots, L$ **do**
7: Process tokens by using $l$-th Transformer block: $\mathrm{T}^l \leftarrow \mathrm{block}(\mathrm{T}^{l-1})$
8: **if** $l \leqq A$ **then** # A-block
9: $\mathrm{X}^l \leftarrow \mathrm{accumulate}(\mathrm{T}^{l-1}, \mathrm{X}^{l-1})$ # Eq. 1
10: **else** # M-block
11: $\mathrm{H}^l \leftarrow \mathrm{accumulate}(\mathrm{T}^{l-1}, \mathrm{X}^{l-1})$ # Eq. 2
12: $\mathrm{X}^l \leftarrow \mathrm{graph_conv}(\mathrm{H}^l, \mathrm{C})$ # Eq. 3
13: $\mathrm{T}^l \leftarrow \mathrm{modulate}(\mathrm{X}^l, \mathrm{T}^l)$ # Eq. 4
14: **end if**
15: **end for**
16: **return** $\mathrm{T}^L$

In Eq. 1, $D$ represents a down-projection function implemented as a linear layer. $D$ compresses each token from $d$-dim. to $d'$-dim. In this paper, $d'$ is set to half of $d$, i.e., 192. The compressed token is added to the token from the preceding A-block, i.e., $\mathbf{x}_i^{l-1}$. At the first A-block, i.e., $l$=1, we use a zero-vector as $\mathbf{x}_i^{l-1}$. After the summation in Eq. 1, $\mathbf{x}_i^l$ is sent to the subsequent A-block. The output from the final (or $A$-th) A-block is passed to the first M-block. Through this accumulation process, all tokens from earlier Transformer blocks can attend to token adaptation in the M-blocks.

### 3.1.3 Modulation block (M-block)

The M-blocks modulate the tokens to adapt them for a downstream task. Identical to an A-block, each M-block first adds the tokens from the frozen Transformer block (i.e., $\mathbf{t}_i^{l-1}$) to the tokens from the preceding M-block (i.e., $\mathbf{x}_i^{l-1}$) by using Eq. 2.

$$\mathbf{h}_i^l = D(\mathbf{t}_i^{l-1}) + \mathbf{x}_i^{l-1} \tag{2}$$

The intermediate token feature $\mathbf{h}_i^l$ is then passed to the graph convolutional layer $G$ defined in Eq. 3.

$$\mathbf{x}_i^l = G(\mathbf{h}_i^l) = \varphi\left( \max_{j \in k\mathrm{NN}(\mathbf{c}_i)} \sigma(\mathbf{h}_i^l, \mathbf{h}_j^l) \right) \tag{3}$$

In Eq. 3, $\sigma$ represents the function which encodes the relation between two token features. We employ an efficient relation function as $\sigma$, which will be detailed later. $k$NN($\mathbf{c}_i$) denotes the set of indices for $k$ patch centers closest to $\mathbf{c}_i$. Note that nearest neighbor search is performed in the 3D coordinate space to effectively capture local geometry. This paper uses $k$=8. The $k$ features produced by $\sigma$ are aggregated by max-pooling and then processed by $\varphi$, which is a linear layer from $d'$-dim. to $d'$-dim.

After graph convolution, $\mathbf{x}_i^l$ is up-projected and added to the token $\mathbf{t}_i^l$. This token modulation process is formulated as follows.

$$\mathbf{t}_i^l \leftarrow U(\mathbf{x}_i^l) + \mathbf{t}_i^l \tag{4}$$

In Eq. 4, $U$ is the up-projection function implemented as a linear layer from $d'$-dim. to $d$-dim. The result of graph convolution, i.e., $\mathbf{x}_i^l$, is also used as the input to the subsequent M-block.

**Efficient EdgeConv:** There are several options for the relation function $\sigma$. This paper adopts the relation function of EdgeConv, a powerful and widely used graph convolution operation. The original EdgeConv encodes the relationship between the feature and its one of the neighbors by using Eq. 5. In Eq. 5, $\|$ denotes vector concatenation and $\mathbf{W} \in \mathbb{R}^{2d' \times d'}$ is the learnable weight matrix for feature transformation.

$$\sigma(\mathbf{h}_i^l, \mathbf{h}_j^l) = (\, \mathbf{h}_i^l \,\|\, (\, \mathbf{h}_j^l - \mathbf{h}_i^l) \,)\mathbf{W} \tag{5}$$

However, a notable drawback of the original EdgeConv implementation lies in its inefficient processing pipeline. That is, the original EdgeConv applies feature transformation by $\mathbf{W}$ to each concatenated feature. Given that $k$ different concatenated features are created for each input feature $\mathbf{h}_i^l$, $nk$ features must be transformed per point cloud. Therefore, both temporal and spatial costs of feature transformation by Eq. 5 is roughly $k$ times greater compared to those of a simple linear layer.

To mitigate this computational burden, we reformulate Eq. 5 to improve efficiency while preserving the expressive power. Specifically, splitting $\mathbf{W}$ into two submatrices $\mathbf{W}_1 \in \mathbb{R}^{d' \times d'}$ and $\mathbf{W}_2 \in \mathbb{R}^{d' \times d'}$ allows us to rewrite Eq. 5 as follows.

$$\begin{aligned} &\mathbf{h}_i^l \mathbf{W}_1 + (\mathbf{h}_j^l - \mathbf{h}_i^l)\mathbf{W}_2 \\ &= \mathbf{h}_i^l (\mathbf{W}_1 - \mathbf{W}_2) + \mathbf{h}_j^l \mathbf{W}_2 \end{aligned} \tag{6}$$

By substituting $\mathbf{W}_1 - \mathbf{W}_2$ with a new matrix $\mathbf{W}' \in \mathbb{R}^{d' \times d'}$, we obtain our efficient EdgeConv operation:

$$G(\mathbf{h}_i^l) = \varphi\left( \max_{j \in k\mathrm{NN}(\mathbf{c}_i)} \mathbf{h}_i^l \mathbf{W}' + \mathbf{h}_j^l \mathbf{W}_2 \right) \tag{7}$$

Eq. 7 shows that the relation between the two features ($\mathbf{h}_i^l$ and its neighbor) can be computed by applying different linear projections to them. Our graph convolution eliminates the need for applying feature transformation to $k$ concatenated vectors. Thus, our graph convolution is $k$ times more computationally efficient than the original EdgeConv.

Note that our efficient EdgeConv is not exactly equivalent to the original EdgeConv since the derivation of Eq. 6 ignores bias terms. Nevertheless, our experiment demonstrates that our graph convolution yields classification accuracy nearly identical to that of the original EdgeConv while reducing computational costs.

#### 3.1.4 Two variants of STAG

We propose two variants of STAG. The first variant, denoted as *STAG-std*, employs the standard configuration with $A$=6 for a backbone having $L$=12 Transformer blocks. This configuration bisects the adaptation module into the same number of blocks: the first six blocks constitute A-blocks, while the subsequent six blocks comprise M-blocks. Each layer type ($D$, $U$, or $G$) shares parameters across all blocks. Note that the parameters of $D$ are shared between A-blocks and M-blocks as well. As a result, STAG-std introduces only 0.43M tunable parameters, representing a substantial reduction compared to existing PEFT-PT methods.

The second variant is a slightly larger STAG, denoted as *STAG-sl*. STAG-sl modifies its architecture by reducing $A$ to 3, thereby increasing the number of M-blocks to 9. STAG-sl also relaxes the parameter sharing constraints. Specifically, parameters are shared only across adjacent three layers. For example, among the nine graph convolution layers ($G$), parameters are shared within three consecutive groups: the first three layers, the middle three layers, and the final three layers. This configuration results in approximately 1M tunable parameters, which align with the number of parameters used in existing PEFT-PT methods.

#### 3.1.5 Reduction in gradient computation

To optimize tunable parameters, their gradients must be computed during fine-tuning. The gradients are computed by taking the partial derivative of the loss (cross-entropy in this paper) with respect to the tunable parameters. Since a DNN is a composite of multiple feature transformation functions, backpropagation applies the chain rule to compute gradients for each tunable parameter. The chain rule requires calculating the gradients for all the neuron activations that exist between the tunable parameter and the loss along the backpropagation path. Therefore, conventional PEFT-PT methods, which have tunable parameters in the upstream of backbone Transformer, require computing gradients for all the backbone activations even if backbone parameters are frozen during fine-tuning.

In contrast, STAG, which has tunable parameters within the side network only, enables the reduction in gradient computations. Fig. 3 illustrates an example of STAG having two A-blocks followed by two M-blocks as well as its backpropagation path. We can observe that the fourth Transformer block is included in the backpropagation path, meaning that its gradients must be computed in order to obtain the gradient for the upstream tunable parameters, e.g., $D$ in the A-block. On the other hand, the first three Transformer blocks in Fig. 3 are not included in the backpropagation path. Therefore, their gradient computation is unnecessary. In general, STAG eliminates the need for gradient computations for the first $A$ Transformer blocks belonging to the A-block and the first one Transformer block of the M-block, resulting in temporally and spatially efficient fine-tuning.

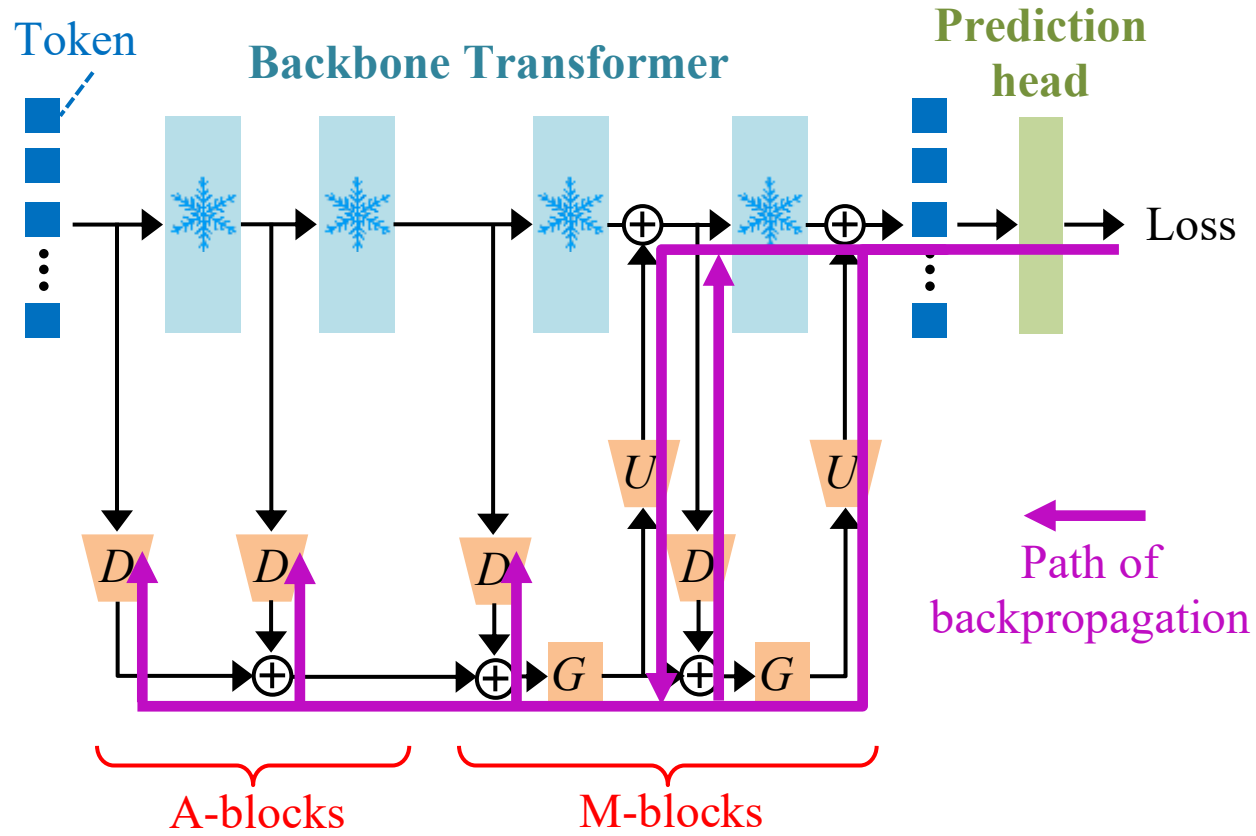

**Fig. 3** In the proposed STAG, the upstream Transformer blocks are not included in the backpropagation path. Therefore, gradient computation for these Transformer blocks is unnecessary during backpropagation, enabling temporally and spatially efficient fine-tuning.

### 3.2 Proposed benchmark: PCC13

We propose PCC13, a comprehensive benchmark for evaluating classification accuracies of diverse 3D point cloud data. Table 1 summarizes the datasets used in PCC13. In the table, "realistic (R)" indicates that the dataset consists of 3D point clouds obtained by scanning real-world objects, while "synthetic (S)" refers to 3D point clouds derived from 3D CAD models. We have composed PCC13 to include publicly available datasets with diverse shape categories, dataset scales, and label granularities.

ScanObjectNN [16] consists of realistic 3D point clouds of indoor objects and is divided into three subsets: obj_bg (objects with background), obj_only (objects without background), and hardest (objects with background and translational/rotational perturbations). OmniObject3D [62] (OmniObject) comprises clean, high-fidelity realistic point clouds classified into diverse object categories. 3DGrocery100 [63] (Grocery100) contains realistic point clouds of food items scanned from a single viewpoint. MVPNet [64] includes diverse realistic point clouds reconstructed from multi-view images. Objaverse-LVIS [65] (Obj. LVIS) consists of both realistic and synthetic point clouds classified into highly diverse and long-tailed categories. ModelNet40 [17] (MN40) comprises synthetic point clouds of rigid objects such as furniture and vehicles derived from 3D CAD models. MCB [66] contains synthetic point clouds of mechanical components. We use the subset B of MCB (denoted as MCB-B) in this paper. SHREC 2015 Non-rigid (SH15NR) [67] consists of synthetic point clouds of animals and objects with varying poses. FG3D [68] comprises synthetic point clouds annotated with fine-grained category labels and includes three subsets: airplane, car, and chair. Since Obj. LVIS and SH15NR do not provide official train/test splits, we randomly divided the entire dataset into training and testing sets. In Obj. LVIS, categories with only one sample were excluded from our experiments. The point clouds in ScanObjectNN contain 2,048 points per object, whereas other datasets have 1,024 points per object.

We evaluate classification accuracy for each dataset listed in Table 1. Fine-tuning is performed separately for each of the 13 datasets. That is, fine-tuning for each dataset starts with the pretrained parameters of the backbone Transformer and the randomly initialized parameters of the STAG side network. The training samples are used for fine-tuning, while the testing samples are used for evaluating classification accuracy. The 13 accuracy values obtained from each dataset are averaged to obtain the overall score for the PCC13 benchmark. Note that the purpose of using PCC13 is not limited to evaluating PEFT-PT methods; it can also be utilized for evaluating 3D point cloud DNN architectures, pre-training algorithms, etc.

## 4. Experiments and their results

### 4.1 Experimental setup

We conduct experiments to comprehensively evaluate the accuracy and efficiency of PEFT-PT algorithms, including the proposed STAG. The experiments utilize the proposed PCC13 benchmark described in Section 3.2.

**Pre-trained Transformers:** We adopt three 3D point cloud Transformers, which have been pre-trained by SSL. Point-MAE [5] is a popular 3D point cloud Transformer that employs the standard Transformer architecture [21]. Point-MAE is pre-trained via the masked autoencoding framework. MaskLRF [28] is a 3D point cloud Transformer that exhibits invariance against SO(3) rotation of input 3D objects. MaskLRF is pre-trained through masked autoencoding of patches, whose orientations are normalized using local reference frame. Uni3D-S [46] is a powerful multi-modal model, whose Transformer is pre-trained via contrastive learning on 3D point clouds, 2D images, and texts. All three pre-trained models have approximately 22M parameters.

**Table 1** Datasets of labeled 3D point clouds used for the PCC13 benchmark.

| Dataset | ScanObjectNN | | | Omni Object | Grocery 100 | MVP Net | Obj. LVIS | MN40 | MCB-B | SH15 NR | FG3D | | |
|---|---|---|---|---|---|---|---|---|---|---|---|---|---|
| | obj_bg | obj_only | hardest | | | | | | | | airplane | car | chair |
| Realistic (R) / synthetic (S) | R | R | R | R | R | R | Both | S | S | S | S | S | S |
| # of categories | 15 | 15 | 15 | 216 | 100 | 180 | 1,145 | 40 | 25 | 50 | 13 | 20 | 33 |
| # of training samples | 2,309 | 2,309 | 11,416 | 4,219 | 66,032 | 70,191 | 31,693 | 9,843 | 14,451 | 600 | 3,441 | 7,010 | 11,124 |
| # of testing samples | 581 | 581 | 2,282 | 1,163 | 21,866 | 17,634 | 14,348 | 2,468 | 3,587 | 600 | 732 | 1,315 | 1,930 |

**Competitors:** STAG is compared against seven fine-tuning methods including two baseline methods and five existing PEFT-PT algorithms. Full fine-tuning is the commonly adopted approach that adjusts all parameters of Tokenizer, Transformer, and prediction head. "Pred. head only" adjusts solely the prediction head parameters while keeping other components frozen. IDPT [10], DAPT [11], Point-PEFT [12], PPT [13], and PointGST [15] are existing PEFT-PT algorithms that combine the adapter tuning and prompt tuning strategies. STAG and the seven competitors use the same prediction head architecture, which is a three-layer MLP adopted by Point-MAE.

**Fine-tuning configurations:** For a systematic evaluation, we use consistent hyper-parameters across all experiments using the PCC13 benchmark. Optimal hyperparameter configurations are expected to vary across different fine-tuning methods, datasets, and backbone Transformers. However, manually tuning hyperparameters for each case is impractical due to time constraints. We thus adopt the de facto standard hyperparameter configurations that are commonly employed for fine-tuning 3D point cloud Transformers. Specifically, we use the optimizer AdamW [69] with an initial learning rate of $5\times10^{-4}$. The learning rate gradually decreases to $1\times10^{-6}$ following a Cosine scheduling. Each pre-trained Transformer is fine-tuned for 300 epochs with a batch size of 32. For data preprocessing, we first normalize the position of a 3D point cloud by translating its gravity center to the origin of the 3D space. The scale of the point cloud is then normalized by fitting its entire shape within a unit sphere. During fine-tuning, we augment each training 3D point cloud by anisotropic scaling and translation. The scaling factors for each axis are randomly sampled from U(0.67, 1.5), while the translation displacements for each axis are sampled from U(–0.2, 0.2).

To evaluate the potential capability of each fine-tuning method, we sample the highest classification accuracy achieved during 300 epochs for each experiment, rather than the accuracy at the 300th epoch. We repeat each experiment three times with different random seeds and report their mean accuracy.

**Hardware configurations:** The experiments were conducted on a PC equipped with an *Intel Core i7-14700KF* CPU, 128GB of main memory, and an *NVIDIA RTX 6000 Ada* GPU with 48GB VRAM.

## 4.2 Experimental results

### 4.2.1 Comparison of accuracy

We first evaluate classification accuracy of STAG and its competitors by using the PCC13 benchmark. Tables 2, 3, and 4 present the classification accuracies on PCC13 using Point-MAE, MaskLRF, and Uni3D-S, respectively. In these tables, cells containing the two highest accuracies are highlighted in green, while cells containing the two lowest accuracies are highlighted in red for each dataset. From Tables 2, 3, and 4, we can observe several consistent patterns.

- Full fine-tuning shows varying degrees of efficacy depending on the pre-trained model and dataset. In some cases, full fine-tuning exhibits the lowest classification accuracy among the nine methods. As discussed in Section 1, this result may be attributed to either overfitting or forgetting knowledge acquired during pre-training. Successful full fine-tuning would require careful selection of hyperparameters such as learning rate for each combination of pre-trained model and dataset.
- Pred. head only consistently shows the lowest accuracy in most cases. Simply appending an MLP to a frozen Transformer does not allow the pre-trained model to be fully adapted to downstream tasks, suggesting the need for token adaptation.
- Among the existing PEFT-PT methods, more recent algorithms, i.e., PPT, and PointGST, demonstrate superior classification accuracy compared to earlier approaches such as IDPT, DAPT, and Point-PEFT. This trend aligns with the experimental results reported in the original papers of PPT and PointGST.
- The proposed STAG, while not always the best, achieves reasonably good results across many datasets, contributing to favorable overall scores. Notably, STAG-sl, a slightly larger variant, exhibits classification accuracy comparable or even superior to the existing PEFT-PT methods.

Despite the simple network architecture, STAG successfully adapts the frozen pre-trained model to downstream tasks. This positive result probably stems from two reasons. The first reason is the complementary interaction between global shape context and local geometry. That is, the backbone Transformer excels at capturing the relations among tokens at a global scale through the self-attention mechanism. In contrast, STAG's graph convolution refines tokens by considering their proximity in the 3D space, thereby making STAG adept at capturing local geometry. These two types of features, i.e., global shape context and local geometry, synergistically complement each other for effective token adaptation. This hypothesis is supported by the fact that STAG achieves favorable results on FG3D, a fine-grained category dataset, where capturing local geometric features as well as global features is crucial.

The second reason is the avoidance of overfitting. The simple network architecture, consisting only of linear and graph convolutional layers, has inherently limited expressive power. Additionally, parameter sharing serves as a strong regularization during fine-tuning. As a result, STAG effectively avoids overfitting and achieves favorable classification accuracy. However, this limited expressiveness can sometimes be disadvantageous, as evidenced by, for example, STAG-std's accuracy on Grocery100 and MVPNet in Table 4. Future work should address the challenge of adjusting STAG's hyperparameters according to the complexity of both the pre-trained model and the dataset for fine-tuning.

**Table 2** Classification accuracies [%] for the PCC13 benchmark (pre-trained model: **Point-MAE** [5])

| Algorithms | ScanObjectNN | | | Omni Object | Grocery 100 | MVP Net | Obj. -LVIS | MN40 | MCB-B | SH15 NR | FG3D | | | Overall |
|---|---|---|---|---|---|---|---|---|---|---|---|---|---|---|
| | obj_bg | obj_only | hardest | | | | | | | | airplane | car | chair | |
| Full fine-tuning | 89.4 | 87.8 | 84.8 | 71.1 | 50.3 | 91.7 | 39.0 | 92.7 | 94.9 | 96.4 | 96.1 | 75.3 | 80.9 | 80.8 |
| Pred. head only | 83.2 | 85.7 | 73.7 | 61.8 | 29.1 | 63.1 | 32.2 | 92.1 | 91.3 | 89.2 | 95.2 | 64.9 | 77.5 | 72.2 |
| IDPT [10] | 90.1 | 88.4 | 84.6 | 69.1 | 46.1 | 87.1 | 39.0 | 93.2 | 94.7 | 97.0 | 96.0 | 75.9 | 81.7 | 80.2 |
| DAPT [11] | 89.8 | 89.2 | 83.7 | 70.1 | 48.8 | 88.3 | 39.5 | 93.2 | 94.6 | 97.4 | 95.5 | 75.5 | 81.7 | 80.6 |
| Point-PEFT [12] | 90.2 | 89.0 | 85.1 | 70.3 | 47.8 | 84.5 | 39.3 | 94.0 | 94.4 | 95.7 | 95.9 | 76.2 | 81.7 | 80.3 |
| PPT [13] | 89.6 | 89.2 | 83.8 | 72.0 | 49.0 | 89.8 | 39.1 | 92.9 | 94.9 | 97.7 | 96.1 | 74.8 | 80.6 | 80.7 |
| PointGST [15] | 89.4 | 89.2 | 84.4 | 70.9 | 49.1 | 86.6 | 40.4 | 93.3 | 94.5 | 97.0 | 96.4 | 76.2 | 81.9 | 80.7 |
| STAG-std (ours) | 91.5 | 89.4 | 85.1 | 71.0 | 48.9 | 88.4 | 40.1 | 93.0 | 94.9 | 99.3 | 96.2 | 76.4 | 82.0 | 81.2 |
| STAG-sl (ours) | 91.3 | 89.0 | 85.7 | 71.4 | 51.6 | 90.5 | 40.7 | 93.1 | 95.0 | 99.2 | 96.2 | 76.7 | 81.9 | 81.7 |

**Table 3** Classification accuracies [%] for the PCC13 benchmark (pre-trained model: **MaskLRF** [28])

| Algorithms | ScanObjectNN | | | Omni Object | Grocery 100 | MVP Net | Obj. -LVIS | MN40 | MCB-B | SH15 NR | FG3D | | | Overall |
|---|---|---|---|---|---|---|---|---|---|---|---|---|---|---|
| | obj_bg | obj_only | hardest | | | | | | | | airplane | car | chair | |
| Full fine-tuning | 91.8 | 89.8 | 86.9 | 74.9 | 48.1 | 93.6 | 38.9 | 90.0 | 95.9 | 100.0 | 95.2 | 74.6 | 80.2 | 81.5 |
| Pred. head only | 87.1 | 85.7 | 77.0 | 71.2 | 32.6 | 83.7 | 36.7 | 88.7 | 95.4 | 100.0 | 95.2 | 72.0 | 74.1 | 76.9 |
| IDPT [10] | 91.5 | 88.0 | 83.5 | 73.9 | 44.1 | 91.3 | 40.2 | 90.7 | 96.0 | 100.0 | 95.7 | 75.3 | 79.6 | 80.7 |
| DAPT [11] | 90.8 | 88.3 | 82.8 | 71.5 | 45.1 | 91.0 | 38.7 | 89.4 | 95.8 | 100.0 | 95.8 | 73.5 | 77.6 | 80.0 |
| Point-PEFT [12] | 92.5 | 88.6 | 82.5 | 73.0 | 45.6 | 90.9 | 39.7 | 91.0 | 95.9 | 100.0 | 95.7 | 75.5 | 79.6 | 80.8 |
| PPT [13] | 92.2 | 89.0 | 85.0 | 74.9 | 47.6 | 92.4 | 39.0 | 90.3 | 96.0 | 100.0 | 95.9 | 75.4 | 79.4 | 81.3 |
| PointGST [15] | 91.5 | 88.6 | 84.6 | 74.3 | 46.6 | 91.2 | 40.0 | 90.1 | 95.8 | 100.0 | 95.9 | 75.7 | 80.1 | 81.1 |
| STAG-std (ours) | 91.7 | 88.8 | 84.1 | 74.6 | 46.5 | 91.0 | 40.3 | 90.4 | 96.0 | 100.0 | 95.6 | 75.7 | 80.1 | 81.1 |
| STAG-sl (ours) | 92.5 | 88.6 | 84.8 | 75.1 | 48.4 | 92.1 | 40.7 | 90.5 | 96.0 | 100.0 | 95.5 | 75.9 | 79.7 | 81.5 |

**Table 4** Classification accuracies [%] for the PCC13 benchmark (pre-trained model: **Uni3D-S** [46])

| Algorithms | ScanObjectNN | | | Omni Object | Grocery 100 | MVP Net | Obj. -LVIS | MN40 | MCB-B | SH15 NR | FG3D | | | Overall |
|---|---|---|---|---|---|---|---|---|---|---|---|---|---|---|
| | obj_bg | obj_only | hardest | | | | | | | | airplane | car | chair | |
| Full fine-tuning | 93.6 | 91.8 | 87.6 | 75.0 | 48.3 | 91.8 | 40.5 | 93.2 | 95.3 | 99.6 | 96.5 | 76.2 | 81.2 | 82.4 |
| Pred. head only | 93.6 | 92.8 | 85.4 | 71.7 | 31.1 | 68.5 | 43.8 | 93.3 | 93.6 | 98.1 | 95.9 | 74.3 | 81.5 | 78.7 |
| IDPT [10] | 93.0 | 91.8 | 87.1 | 72.3 | 45.4 | 85.9 | 41.4 | 93.5 | 94.8 | 95.1 | 97.1 | 77.7 | 82.6 | 81.4 |
| DAPT [11] | 94.0 | 92.7 | 88.6 | 74.5 | 47.4 | 90.7 | 45.6 | 93.8 | 95.3 | 99.1 | 97.1 | 76.9 | 82.1 | 82.9 |
| Point-PEFT [12] | 94.6 | 93.6 | 89.7 | 75.0 | 49.3 | 88.4 | 46.4 | 94.4 | 95.1 | 98.3 | 96.9 | 77.6 | 82.3 | 83.2 |
| PPT [13] | 94.9 | 92.5 | 89.1 | 76.2 | 50.6 | 92.4 | 46.3 | 93.6 | 95.5 | 99.6 | 97.0 | 77.6 | 82.7 | 83.7 |
| PointGST [15] | 94.9 | 92.6 | 89.3 | 76.6 | 50.4 | 90.5 | 47.9 | 93.9 | 95.3 | 99.9 | 96.7 | 77.8 | 82.8 | 83.7 |
| STAG-std (ours) | 94.8 | 92.9 | 88.7 | 75.7 | 46.4 | 87.6 | 45.6 | 94.2 | 95.3 | 99.9 | 96.8 | 78.1 | 83.0 | 83.0 |
| STAG-sl (ours) | 94.8 | 92.8 | 89.1 | 76.8 | 50.3 | 90.4 | 46.5 | 94.2 | 95.5 | 99.9 | 97.1 | 78.8 | 83.5 | 83.8 |

#### 4.2.2 Comparison of efficiency

This subsection evaluates fine-tuning methods not only in terms of parameter efficiency but also temporal and spatial efficiency. Table 5 compares the efficiency of the fine-tuning methods. An important finding is that the existing PEFT-PT methods are not necessarily temporally and spatially efficient; some methods actually show inferior computational efficiency compared to full fine-tuning. Among the methods in Table 5, the prediction head only approach shows the best computational efficiency, which is expected given its framework. We exclude pred. head only from subsequent discussion due to its accuracy limitations, as evidenced in Section 4.2.1.

**Parameter efficiency:** the primary objective of PEFT is to reduce the number of tunable parameters during fine-tuning. Accordingly, Table 5 shows that all PEFT-PT methods demonstrate significantly improved parameter efficiency compared to full fine-tuning. Table 5 also reveals that our proposed STAG-std achieves the lowest number of tunable parameters (0.43M), requiring only 2% of the parameters for full fine-tuning. This high parameter efficiency stems from STAG-std's simple network architecture and parameter sharing framework. Even STAG-sl, which relaxes parameter sharing constraints, maintains competitive efficiency to several existing PEFT-PT methods with approximately 1M tunable parameters.

**Temporal efficiency for fine-tuning:** We use two metrics: the number of floating-point operations (FLOPs) and actual computation time per epoch. FLOPs are counted separately for inference (denoted as forward) and backpropagation (denoted as backward). As shown in Table 5, STAG-std and STAG-sl maintain inference costs similar to full fine-tuning while significantly reducing backpropagation costs. This reduction is achieved by eliminating gradient computation in the earlier Transformer blocks. The gradient calculation terminates at the sixth block for STAG-std and the third block for STAG-sl. In contrast, existing PEFT-PT methods require gradient computation through to the first Transformer block or its preceding Tokenizer. The reduced gradient calculation allows STAG to accelerate fine-tuning. As shown in the column “Time per epoch” in Table 5, STAG-std runs approximately 1.7 times faster than full fine-tuning and 1.4 times faster than DAPT, which is the fastest among the existing PEFT-PT methods. As we demonstrate in the next subsection, using efficient EdgeConv also contributes to the acceleration.

**Table 5** Efficiency comparison of fine-tuning methods using the ScanObjectNN obj_bg dataset and batch size of 32.

| Algorithms | Complexity for fine-tuning | | | | | Inference time per batch [s] |
|---|---|---|---|---|---|---|
| | # of tuned parameters | Forward [GFLOPs] | Backward [GFLOPs] | Time per epoch [s] | VRAM [GB] | |
| Full fine-tuning | 22.09M | **314** | 629 | 4.29 | 6.1 | **0.032** |
| Pred. head only | 0.27M | 314 | 0.02 | 1.70 | 0.9 | 0.032 |
| IDPT [10] | 1.70M | 464 | 492 | 5.43 | 6.6 | 0.062 |
| DAPT [11] | 1.09M | 328 | 214 | 3.57 | 4.7 | 0.033 |
| Point-PEFT [12] | 0.77M | 501 | 561 | 13.66 | 13.2 | 0.104 |
| PPT [13] | 1.04M | 724 | 859 | 9.42 | 13.5 | 0.063 |
| PointGST [15] | 0.62M | 319 | 203 | 5.59 | 3.6 | 0.066 |
| STAG-std (ours) | **0.43M** | 331 | **110** | **2.59** | **2.0** | 0.035 |
| STAG-sl (ours) | 1.02M | 335 | 169 | 3.10 | 3.0 | 0.037 |

The superior fine-tuning speed of the proposed STAG would become more pronounced on larger-scale datasets. For instance, consider fine-tuning using the Objaverse 1.0 dataset [65], which contains 800K 3D objects (346 times larger than ScanObjectNN obj_bg). Full fine-tuning would require approximately 5 days of training time. In contrast, STAG-std, which is 1.7 times faster than full fine-tuning, would require approximately 3 days, achieving a significant reduction of 2 days in training time. Although we cannot conduct experiments since Objaverse 1.0 lacks class labels, the high efficiency of STAG will prove valuable when large-scale classification datasets emerge in the future.

**Spatial efficiency for fine-tuning:** Table 5 also compares GPU memory consumption during fine-tuning. Generally, a GPU during fine-tuning must store: (1) DNN parameters, (2) optimizer states (e.g., moving average and variance for each tunable parameter for AdamW), (3) gradients for parameters, and (4) forward activations saved for gradient computation. STAG can reduce all these information by decreasing the numbers of tunable parameters and gradient calculations. As a result, STAG-std and STAG-sl require only 2GB and 3GB of VRAM respectively. In particular, STAG-std achieves a 40% memory reduction compared to PointGST, which is the most memory-efficient among the existing methods. Fig. 4 plots GPU memory consumption against batch size. In the figure, the absence of data point indicates the occurrence of a GPU out-of-memory error. All methods show linear memory growth with batch size (note the logarithmic scale for the horizontal axis). STAG exhibits the most gradual memory increase and is the only method capable of fine-tuning with a batch size of 512.

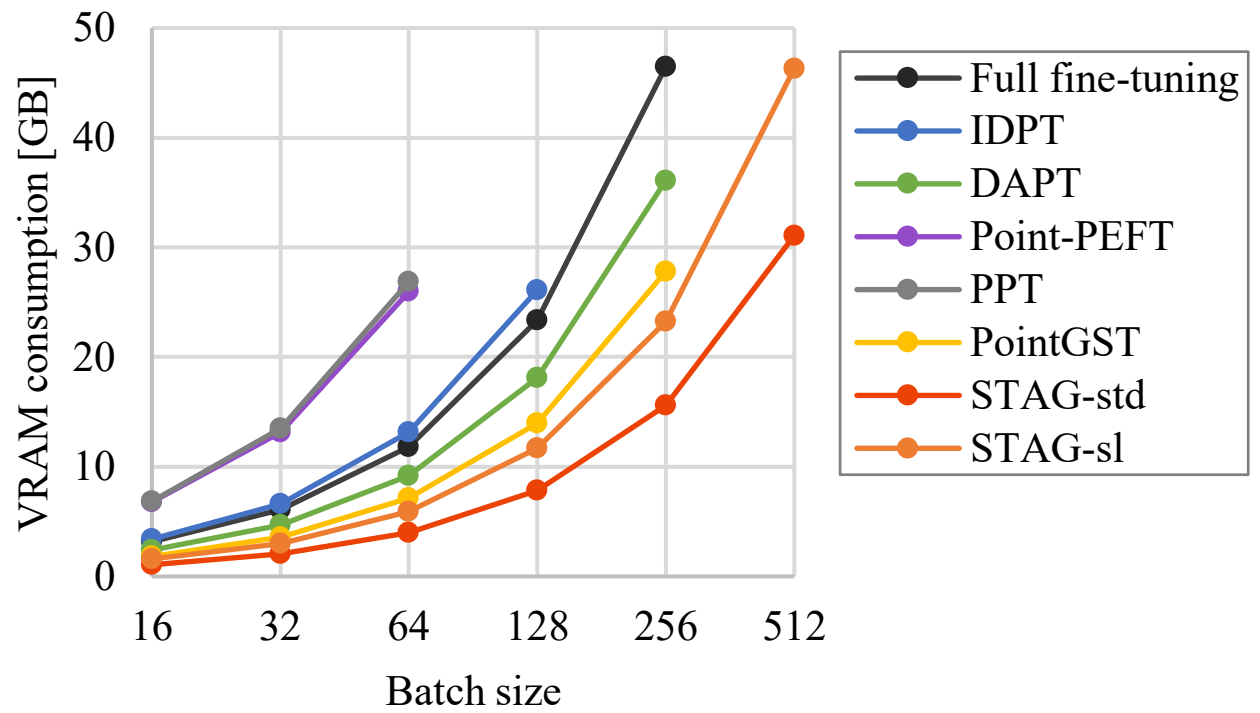

**Fig. 4** GPU memory footprint plotted against batch size.

**Temporal efficiency for inference:** Inference time is also important for real world deployment after fine-tuning. Table 5 shows that full fine-tuning and “Pred. head only” offer the fastest inference speeds. This is an expected result, as these methods do not involve additional adaptation modules. Among the PEFT-PT methods, DAPT is the fastest due to its use of a simple adapter module based on fully connected layers. Following DAPT, our proposed STAG-std and STAG-sl are the next most efficient, with computation times increasing by only 9% and 13%, respectively, compared to full fine-tuning. Despite the complex operation (i.e., graph convolution) of STAG, the increase in latency is small; this is attributed to Efficient EdgeConv proposed in Section 3.1.3. In contrast, other existing PEFT-PT methods, i.e., IDPT, Point-PEFT, PPT, and PointGST, exhibit double or more the inference time of full fine-tuning, making them less time-efficient for real world deployment.

To summarize the experiments in this subsection, the proposed STAG achieves equivalent or fewer tunable parameters, faster training times, and reduced GPU memory consumption compared to existing PEFT-PT algorithms. This result validates our successful development of a temporally and spatially efficient PEFT framework for 3D point cloud Transformers, which is our primary goal of this paper.

#### 4.2.3 In-depth evaluation of STAG

This subsection validates the design choices of STAG. We conduct experiments using STAG-std with Point-MAE as the pre-trained model on four datasets: ScanObjectNN obj_bg, OmniObject, MVPNet, and Obj. LVIS.

**Effectiveness of parameter sharing:** Table 6 demonstrates the impact of parameter sharing on the number of tunable parameters and classification accuracy. Disabling parameter sharing means that all A-blocks and M-blocks use distinct parameters during fine-tuning. As shown in Table 6, enabling parameter sharing reduces the number of tunable parameters to approximately one-fifth, indicating significant improvement in parameter efficiency. Interestingly, enabling parameter sharing shows a small accuracy drop (~0.2%) on three datasets except for MVPNet. We attribute the nearly 2% accuracy drop in MVPNet, which has the largest number of training samples among PCC13, to underfitting due to the reduced parameters.

**Table 6** Effectiveness of parameter sharing.

| Share parameters? | # of tuned parameters | Classification accuracy [%] | | | |
|---|---|---|---|---|---|
| | | obj_bg | OmniObject | MVPNet | Obj. LVIS |
| No | 2.17M | 91.7 | 71.2 | 90.2 | 40.2 |
| Yes | 0.43M | 91.5 | 71.0 | 88.4 | 40.1 |

**Influence of number of A-blocks and M-blocks:** Table 7 shows how the number of the blocks affects computational efficiency and classification accuracy. We vary the value of $A$ to control the allocation between A-blocks and B-blocks. In terms of efficiency, Table 7 shows that both computational time and memory consumption grow as the number of M-blocks increases. The increased computational cost with more M-blocks stems from the overheads by graph convolutions during inference and gradient calculations during backpropagation. Due to the parameter sharing framework, the number of tunable parameters remains constant regardless of the value of $A$. That is, the parameters of the three functions, i.e., down projection $D$, up projection $U$, and graph convolution $G$ are shared across all A-blocks and M-blocks, leading to the fixed number of tunable parameters. Regarding accuracy, three datasets except for ScanObjectNN obj_bg show a trend where using more M-blocks leads to better classification accuracy. This trend is probably because applying more token modulation allows better adaptation to the three datasets, which contain a large number of categories and/or training samples. On the other hand, using too many M-blocks may have led to overfitting for a small-scale dataset such as ScanObjectNN obj_bg. Considering the balance between computational efficiency and classification accuracy, the configuration of $A$=6 in STAG-std appears to be a reasonable choice.

**Effectiveness of efficient graph convolution:** Table 8 compares various operations used as the token refinement function $G$ within M-blocks. In the table, max-pooling simply aggregates neighboring tokens through max-pooling without feature transformation by a linear layer. Scalar self-attention refines neighboring tokens using the original self-attention [21] that computes attention scores for each pair of tokens. Vector self-attention [70] performs token refinement using attention scores computed per feature channel. Simple graph conv. [71] is the most basic form of graph convolution, which applies a linear transformation before local max-pooling. Original EdgeConv [20] performs graph convolution using the relation function shown in Eq. 5.

Table 8 demonstrates that the proposed efficient EdgeConv successfully achieves both low computational cost and high classification accuracy. Compared to the original EdgeConv, our efficient version reduces the temporal cost by approximately 10% and memory cost by about 30%. The reformulation that eliminates the need for concatenated vectors contributes to reducing the overhead in EdgeConv computation. Table 8 also shows that self-attention produces inferior classification accuracy compared to graph convolution. This is probably because self-attention's high expressiveness leads to overfitting to downstream task datasets.

**Table 7** Influence of the number of A-blocks and M-blocks in STAG.

| # of A-blocks (value for $A$) | # of M-blocks | # of tuned parameters | Time per epoch [s] | VRAM [GB] | Classification accuracy [%] | | | |
|---|---|---|---|---|---|---|---|---|
| | | | | | obj_bg | OmniObject | MVPNet | Obj. LVIS |
| 11 | 1 | 0.43M | 2.18 | 1.1 | 91.0 | 71.0 | 86.5 | 39.9 |
| 10 | 2 | 0.43M | 2.29 | 1.2 | 91.2 | 71.0 | 87.7 | 39.9 |
| 8 | 4 | 0.43M | 2.73 | 1.5 | 91.2 | 71.0 | 87.6 | 39.9 |
| 6 | 6 | 0.43M | 2.59 | 2.0 | **91.5** | 71.1 | 88.4 | 40.1 |
| 4 | 8 | 0.43M | 3.24 | 2.7 | 91.2 | 71.0 | 88.7 | 40.2 |
| 2 | 10 | 0.43M | 3.47 | 3.3 | 91.3 | 71.2 | 88.7 | 40.3 |
| 0 | 12 | 0.43M | 3.63 | 3.9 | 91.0 | **71.3** | **88.8** | **40.4** |

**Table 8** Comparison of the token refinement function $G$ in STAG.

| Token refinement functions | # of tuned parameters | Time per epoch [s] | VRAM [GB] | Classification accuracy [%] | | | |
|---|---|---|---|---|---|---|---|
| | | | | obj_bg | OmniObject | MVPNet | Obj. LVIS |
| Max-pooling | 0.35M | 2.57 | 1.9 | 90.5 | 70.6 | 87.7 | 39.9 |
| Scalar self-attention | 0.46M | 3.01 | 2.4 | 87.1 | 66.6 | 81.2 | 35.4 |
| Vector self-attention | 0.54M | 3.10 | 3.6 | 91.1 | 69.8 | 69.6 | 39.1 |
| Simple graph conv. | 0.39M | 2.58 | 2.0 | 90.0 | 70.4 | 87.5 | 39.4 |
| Original EdgeConv | 0.43M | 2.93 | 2.9 | **91.5** | **71.1** | 88.3 | 40.0 |
| Efficient EdgeConv (ours) | 0.43M | 2.59 | 2.0 | **91.5** | **71.1** | **88.4** | **40.1** |

**Influence of neighborhood size:** Table 9 shows the impact of the number of neighbors $k$ for neighborhood graph, which is a crucial hyperparameter for graph convolution. Across all four datasets, classification accuracy improves as $k$ increases from 1 to 8, demonstrating the effectiveness of graph convolution. However, excessively large neighborhood sizes deteriorate classification accuracy. This is probably because, as discussed in Section 4.2.1, the synergistic effect between global shape context and local geometry is lost. Specifically, using large $k$ prevents STAG from achieving effective token adaptation because STAG captures global feature rather than local geometry of an input 3D point cloud.

**Table 9** Influence of neighborhood size for graph convolution.

| Neighborhood size $k$ | Classification accuracy [%] | | | |
|---|---|---|---|---|
| | obj_bg | OmniObject | MVPNet | Obj. LVIS |
| 1 | 88.8 | 70.2 | 82.8 | 38.8 |
| 2 | 89.3 | 70.4 | 83.9 | 39.0 |
| 4 | 90.3 | 71.0 | 87.0 | 39.6 |
| 8 | **91.5** | **71.1** | **88.4** | **40.1** |
| 16 | 91.0 | 70.9 | 87.5 | 39.8 |
| 32 | 90.4 | 69.8 | 87.1 | 39.2 |
| 64 | 89.4 | 68.9 | 85.7 | 38.3 |

### 4.2.4 Evaluation on parts segmentation

Thus far, we have confirmed the efficacy of the proposed STAG only under the point cloud classification scenario. However, the capability of accurate analysis in various tasks is crucial for practical fine-tuning algorithms. This subsection evaluates PEFT-PT methods on the parts segmentation task. We employ the ShapeNetPart dataset [75] and evaluate accuracy of parts segmentation by using Class-mIoU (C-mIoU) and Instance-mIoU (I-mIoU). For all fine-tuning methods used in the experiments, we append the decoder DNN employed in Point-MAE [5] after the backbone Transformer. During fine-tuning, we optimize backbone parameters according to each fine-tuning method along with approximately 5M parameters within the decoder. The increases in the number of tunable parameters and computational cost brought by the decoder are identical across all fine-tuning methods. Hence, this subsection compares segmentation accuracy only. The efficiency differences in the backbone are demonstrated in Section 4.2.2. We employ the AdamW optimizer with an initial learning rate of 0.0001 and cosine scheduling. The batch size is set to 32, and random anisotropic scaling followed by random translation is used for data augmentation. We conduct each experiment three times with different random seeds, and their average accuracy is reported. For PointGST, we cite the accuracy values from the original paper [15] since our implementation suffered from training instability.

**Table 10** Comparison of accuracies [%] for parts segmentation task using the ShapeNetPart dataset. Among the PEFT algorithms, best-performing accuracy is shown in **bold**, and second-best accuracy is underlined.

| Algorithms | Pre-trained models | | | | | |
|---|---|---|---|---|---|---|
| | Point-MAE [5] | | MaskLRF [28] | | Uni3D-S [46] | |
| | C-mIoU | I-mIoU | C-mIoU | I-mIoU | C-mIoU | I-mIoU |
| Full fine-tuning | 84.6 | 85.9 | 83.2 | 85.3 | 84.6 | 86.0 |
| Pred. head only | 83.4 | 85.0 | 80.4 | 83.1 | 82.6 | 84.7 |
| IDPT [10] | 83.7 | 85.6 | 80.8 | 83.4 | 83.2 | 85.3 |
| DAPT [11] | 83.3 | 85.4 | 81.0 | 83.5 | 83.4 | 85.3 |
| Point-PEFT [12] | 83.5 | 85.3 | 80.2 | 83.3 | 83.2 | 85.1 |
| PPT [13] | **84.1** | 85.5 | **82.2** | **84.6** | 83.2 | 85.4 |
| PointGST [15] | 83.8 | **85.8** | — | — | — | — |
| STAG-std (ours) | 83.8 | 85.5 | 81.7 | 84.2 | **83.7** | 85.5 |
| STAG-sl (ours) | 84.0 | 85.6 | **82.2** | 84.5 | **83.7** | **85.6** |

Table 10 shows segmentation accuracies for combinations of three pre-trained models and nine fine-tuning methods. The proposed STAG, particularly the slightly larger variant STAG-sl, demonstrates accuracy comparable to or superior to existing PEFT-PT algorithms. As discussed in Section 4.2.1, we attributed the effectiveness of STAG to the synergy between shape context captured by the Transformer and local geometry captured by graph convolutions in the side network. In the parts segmentation task, local shape features serve as important cues for partitioning a 3D object into multiple semantic parts. Therefore, it is considered that the side network of STAG successfully adapts local shape features, leading to accurate parts segmentation.

## 5. Conclusion and future work

This paper proposed Side Token Adaptation on a neighborhood Graph (STAG), a novel PEFT algorithm for 3D point cloud Transformers (PEFT-PT) that achieves both temporal and spatial efficiency. STAG is the first PEFT-PT algorithm that employs side-tuning, with its core idea being token adaptation via a graph convolutional side network. STAG improves its efficiency by incorporating the parameter sharing framework and efficient graph convolution operator. The adaptation module of STAG operates independently of the backbone Transformer, making STAG versatile and applicable to various 3D point cloud Transformer architectures. In addition, this paper introduced the Point Cloud Classification 13 (PCC13) benchmark to enable comprehensive evaluation using diverse 3D point cloud data.

The extensive experiments using PCC13 demonstrated STAG's effectiveness in two key aspects:

- STAG achieved classification accuracy comparable to or better than existing PEFT-PT algorithms. The combination of a frozen backbone Transformer and STAG effectively enhances global features extracted by self-attention with local geometric features obtained via graph convolution.
- STAG achieved superior temporal and spatial efficiency compared to existing PEFT-PT algorithms. The three key innovations contribute to the enhanced efficiency of STAG: token modulation applied only in the latter part of the backbone, parameter sharing framework, and efficient graph convolution operator.

Future work includes deepening research on PEFT-PT. For example:

- **Further improving computational efficiency:** As STAG is partially independent of the backbone Transformer, gradient computation for the latter half of the backbone remains necessary. Existing side tuning approaches [58, 59], whose side network is completely independent of the backbone, represent a promising direction for improving efficiency.
- **Adaptive scaling of STAG:** Our evaluation using PCC13 revealed that STAG (particularly the smaller variant STAG-std) sometimes suffers from underfitting on certain datasets. The practicality of STAG could be enhanced by automatically adjusting its hyperparameters based on dataset complexity such as number of semantic categories and training samples.
- **Evaluation on diverse tasks:** This paper evaluated PEFT-PT methods on a 3D point cloud classification task. Future research should explore performance on other downstream tasks including semantic segmentation, detection, retrieval, and few-shot classification.